\documentclass[10pt, a4paper]{article}

\usepackage{lrec-coling2024} % this is the new style
\usepackage{amsfonts}
\usepackage{cuted}
\usepackage{caption}
\usepackage{subcaption}
\usepackage{mathtools}
\usepackage{array}
\newcolumntype{P}[1]{>{\centering\arraybackslash}p{#1}}

\title{Jetsons at FinNLP 2024: Towards Understanding the ESG Impact of a News Article using Transformer-based Models}

\name{Parag Pravin Dakle, {\bf \large Alolika Gon}, {\bf \large Sihan Zha}, {\bf \large Liang Wang}, \\ {\bf \large SaiKrishna Rallabandi}, {\bf \large and Preethi Raghavan}}

\address{Fidelity Investments, AI Center of Excellence \\
         \{firstname.lastname\}@fmr.com\\}

\abstract{
In this paper, we describe the different approaches explored by the Jetsons team for the Multi-Lingual ESG Impact Duration Inference (ML-ESG-3) shared task. The shared task focuses on predicting the duration and type of the ESG impact of a news article. The shared task dataset consists of 2,059 news titles and articles in English, French, Korean, and Japanese languages. For the impact duration classification task, we fine-tuned XLM-RoBERTa with a custom fine-tuning strategy and using self-training and DeBERTa-v3 using only English translations. These models individually ranked first on the leaderboard for Korean and Japanese and in an ensemble for the English language, respectively. For the impact type classification task, our XLM-RoBERTa model fine-tuned using a custom fine-tuning strategy ranked first for the English language.
 \\ \newline  }

\begin{document}

\maketitleabstract

\section{Introduction}

ESG (environment, social, and governance) related news can impact the performance and reputation of companies, investors, and regulators. One of the key challenges in ESG impact assessment is to estimate the duration of the ESG impact of a news article \cite{tseng2023dynamicesg}. Different news articles may have different levels of salience, credibility, and relevance for different stakeholders and thus may have different effects on their behavior and outcomes. The LREC-COLING shared task \cite{chen2024MLESG-3} presents a multi-lingual impact duration and level classification task based on news articles. 
 
 %Existing methods for ESG impact assessment often rely on subjective judgments, historical data, or arbitrary time frames, which may not capture the true and dynamic nature of ESG impact. 
 % According to a recent study, ESG news can affect stock returns, investor sentiment, and regulatory actions for up to five years after publication. ESG stands for environmental, social, and governance, and it refers to the criteria that measure an organization's sustainability and ethical impact. ESG factors include climate change, human rights, diversity, corruption, and board independence. ESG is increasingly important for stakeholders who want to align their values with their investments, reduce risks, and enhance long-term returns. However, ESG is complex and dynamic, requiring reliable and consistent data, analysis, and reporting. 

We approach the shared task using the following strategies - (1) Traditional NLP techniques like TF-IDF with logistic regression, SVM \cite{cortes1995support}, and Random Forest classifiers, (2) De-noising the data to evaluate the impact of removing noisy or less informative samples, (3) Fine-tuning multilingual BERT-style models on individual language and entire dataset, (4) Complementing direct fine-tuning for impact duration with self-training using additional English and French ESG articles, (5) Translating all articles to English to simplify the impact duration task, and (6) Creating an ensemble of the best models for the impact duration task.
 % The work presented in this paper aims to develop a novel method for predicting the duration of the ESG impact of a news article. The method's input is the text of the news article, and the output is one of three classes: less than 2 years, 2 to 5 years, and more than 5 years. The method is based on natural language processing and machine learning techniques, and it uses a large and diverse dataset of ESG news articles and their impact indicators. The expected outcome of this project is to provide a more accurate, objective, and scalable way of measuring and comparing the ESG impact of news articles across different domains and contexts.

\section{Related Work}

There has been an increased focus on evaluating the nonfinancial activities of a company, which is typically encapsulated under the title of ESG. \citet{exploring_trends_in_ESG} show that the various topics included in ESG have gradually evolved. \citet{esg_in_finance_topic_analysis} perform a similar analysis across 11 sectors and show that the best ESG-performing financial institutions are actively committed to the code of best practices in governance. Language Modeling and NLP techniques have been the de facto approaches toward automating the estimation of ESG ratings. 

\textbf{Embeddings for ESG Classification}: \citet{raman2020embeddingbasedESG} investigated employing embeddings from pre-trained language models for classifying sentences relevant to the ESG domain. \citet{esgbert} pre-trained a BERT model on ESG-related text, demonstrating improvements in classification tasks related to ESG factors.

\textbf{Fine-tuning}: \citet{nugent2021detecting} fine-tuned an English BERT-style model specifically for ESG document classification. They explored data generation as an augmentation strategy, enhancing model performance. \cite{jorgensen2021mdapt, jorgensen2023multifin} extended the concept of pre-training language models from financial text to multilingual text and evaluated sentence classification and financial topic classification. 

\section{Data}
\begin{figure*}[t!]
    \centering
    \begin{subfigure}{0.31\linewidth}
        \centering
        \includegraphics[width=\textwidth]{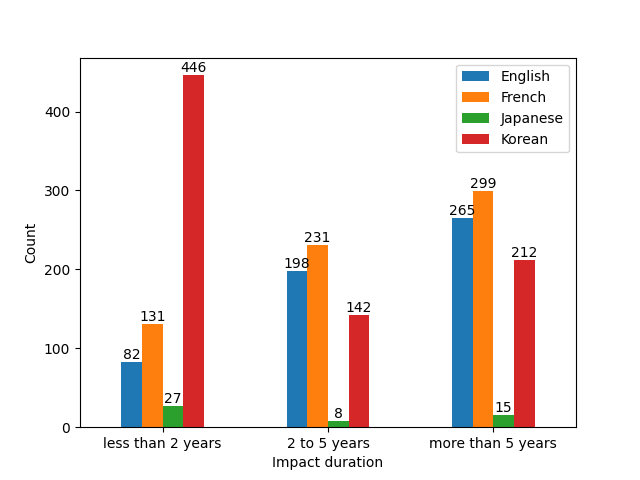}        
        \caption{Impact Duration training data}
        \label{fig:orig_dist_impact_duration}
    \end{subfigure}
    \hfill
    \begin{subfigure}{0.31\linewidth}
        \centering
        \includegraphics[width=\textwidth]{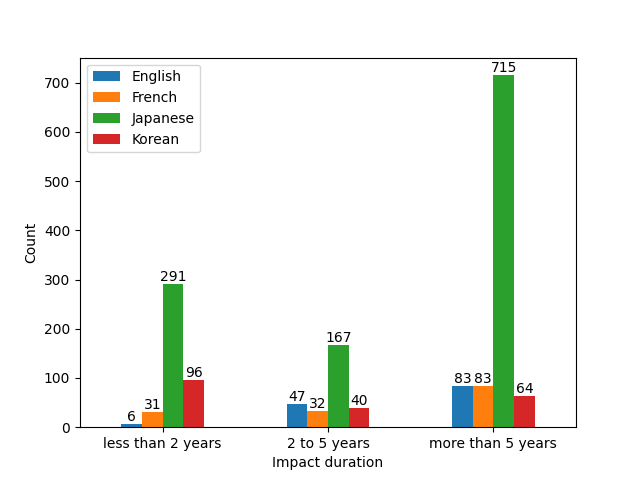}
        \caption{Impact duration test data}
        \label{fig:test_gold_impact_duration}
    \end{subfigure}
    \hfill
    \begin{subfigure}{0.31\linewidth}
        \centering
        \includegraphics[width=\textwidth]{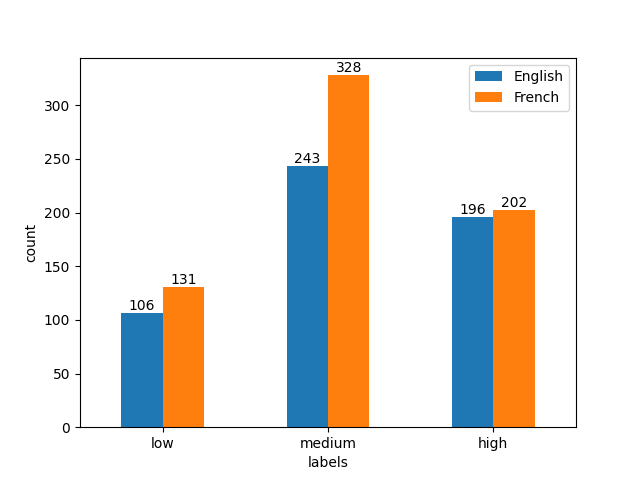}        
        \caption{Impact level training data}
        \label{fig:orig_dist_impact_level}
    \end{subfigure}    
    \caption{Data distribution across classes}
    \label{fig:orig_dist}
\end{figure*}

The training dataset consists of 2,059 news articles in four languages: 545 English(en), 661 French(fr), 800 Korean(kr), and 53 Japanese(jp) articles. Each article has an associated title and the main content. News articles in all four languages are annotated with impact duration labels: less than 2 years, 2 to 5 years, and more than 5 years\footnote{The label names are different for some languages}. The distribution across the 3 impact duration classes is highly skewed, as shown in Figure \ref{fig:orig_dist_impact_duration}. The French and English articles are also annotated with `low', `medium', or `high' impact level classes (Figure \ref{fig:orig_dist_impact_level}). The Korean dataset also contains impact type annotations with the following classes - opportunity, risk, cannot distinguish. This paper does not focus on the Korean impact type classification task. 
% general description, distribution per language

Additionally, 31 and 24 duplicates were encountered while pre-processing the data for the Korean and French training data, respectively. We ignore all duplicates with the same class labels, but for 17 of the 24 French duplicates, we randomly select one of the duplicates to be part of the training dataset. This data is split into 10 parts for 10-fold cross-validation with consistent data distribution across all folds in the training and validation sets. The training and validation set lengths were about 1800 and 200, respectively. Lastly, we also found that the test set for Korean contains 1 duplicate, and for Japanese, it contains 19 duplicates and 327 samples with no class label.

\section{Impact Duration Task}

\subsection{Traditional NLP Methods}
% To explore the prediction tasks, we explore both traditional TF-IDF model and the LLM fine-tuning models.
\subsubsection{Baseline Model}
The small size of the dataset and high frequency of ESG keywords motivated us to evaluate naive TF-IDF classifier models as a traditional NLP baseline.

We consider logistic regression, random forest, and SVM as our baseline models and adopt 10-fold cross-validation for model training and evaluation. To enable the hyperparameters tuning for those baseline models, we further divide the 10-fold training set into train/val with a ratio of 80/20. 
%This setup provides us with 10 folds of train/val/test according to a ratio of 72/18/10.
% which supports multilingual tasks evidenced by its LLMs performance such as mt-5\citep{xue2020mt5}
\citet{wangperawong2022multilingual} show that using a single vocabulary for all languages and subword tokenization greatly improves the classification results. We use SentencePiece\footnote{https://github.com/google/sentencepiece} for multilingual tokenization. We convert the obtained tokens to lowercase and compute TF-IDF statistics with filters of maximum frequency(0.7). We tune the penalty parameters $C$ for SVM and logistic regression, and \textit{number of trees, maximum depth parameter, and minimum sample of internal nodes} parameters for the Random Forest (RF) model. The averaged statistics in percentage from the 10-fold testing set are reported in Table~\ref{tab:baselinef1}. The RF model does a good job predicting the impact duration with large variation for the Japanese due to the smaller dataset.

% \begin{strip}
%     \centering
%     \includegraphics[width = \textwidth]{figures/weighted_f1.png}
%     \label{fig:baselinef1}
%     \captionof{figure}{Impact Length: Weighted F1 across 10-folds for baseline models}
% \end{strip}

\begin{table*}[t]
    \centering
    \begin{tabular}{lcccccc}
        \hline
        \textbf{Model} & \textbf{Setting} & \textbf{overall} & \textbf{kr}    & \textbf{jp}    & \textbf{en}    & \textbf{fr} \\ 
        \hline
        Logistic & Normal & 53.85 & 63.96 & 64.78 & 53.47 & 50.44 \\
        SVM & Normal & 53.61 & 64.02 & 72.59 & 51.34 & 49.27 \\
        Random Forest (RF) & Normal & 58.23 & 71.13 & 61.45 & 61.45 & 58.67 \\ \hline
        Logistic & De-noised & 54.61 & 66.82 & 72.49 & 52.36 & 52.41 \\
        SVM & De-noised & 55.96 & 68.42 & 73.13 & 54.98 & 52.79 \\
        Random Forest (RF) & De-noised & 58.60 & 71.03 & 70.00 & 62.98 & 60.35 \\
        \hline
    \end{tabular}
    \caption{Impact Duration: Weighted F1 averaged across 10-folds for baseline models}
    \label{tab:baselinef1}
\end{table*}

\subsubsection{Learning with De-noised Labels}
Although the impact duration of the ESG news has been cross-validated with agreement statistics across different annotators, it is sometimes challenging to classify an ESG event into less than 2 years, 2 to 5 years, and more than 5 years window. For example, \textit{The new agreements bring Verizon’s projected renewable energy capacity to more than 3GW, enough to power more than 707,000 homes for a year and position the company to meet its goal to source or generate renewable energy equivalent to 50\% of its total annual electricity consumption by 2025.} This article was annotated to be `2 to 5 years' probably due to the knowledge of the time difference between 2025 and the year of the annotation. However, the text clearly indicates a time window of one year, which could or should be annotated as "less than 2 years". The ground truth label of this event can hence be ambiguous. \citet{brodley1999identifying} demonstrated that direct training based on the "mislabeled" data generates less desirable models than training with less but de-noised data. Following a similar idea in \citet{Wang2023}, we explored a data quality model to score each text-label pair.

Using the RF baseline model fine-tuned on TF-IDF tokens, we evaluate on each of the 10-fold testing sets to obtain the confidence of the prediction $\mathbb{P}$ and the label of the prediction $\hat{\mathbb{Y}}$. Then comparing against the annotation from the ground truth labels $\mathbb{Y}$, we compute a quality score $\mathbb{Q}: \mathbb{Y} \times \hat{\mathbb{Y}} \times \mathbb{P} \rightarrow [-1,1]$ using $\mathbb{Q} (\mathbb{P}, \mathbb{Y}, \hat{\mathbb{Y}})= - \mathbb{P}$ if $\mathbb{Y} \neq \hat{\mathbb{Y}}$ and $\mathbb{Q} (\mathbb{P}, \mathbb{Y}, \hat{\mathbb{Y}})= + \mathbb{P}$ if  $\mathbb{Y} = \hat{\mathbb{Y}}$.
% \begin{equation}
% \mathbb{Q} (\mathbb{P}, \mathbb{Y}, \hat{\mathbb{Y}})= 
% \begin{dcases*}
% - \mathbb{P}
%     & if  $\mathbb{Y} \neq \hat{\mathbb{Y}}$\\
% + \mathbb{P}
%    & if $\mathbb{Y} = \hat{\mathbb{Y}}$
% \end{dcases*}
% \end{equation}
Hence, a high-quality score $\mathbb{Q}$ would indicate agreement and high confidence between the predicted labels and the actual labels, whereas a low-quality score $\mathbb{Q}$ indicates agreement with high confidence. Computing on each of the 10-fold testing sets, we obtained the quality score $\mathbb{Q}$ for the entire 2,059 observations, based upon which we delete x\% of the data that are potentially of low quality/agreement. Through our evaluation, we have found that deleting 10\% of the original data provides a decent improvement with the weighted F1 score shown in Table~\ref{tab:baselinef1}. This indicates a certain level of noisiness within the duration labels. 
% However, this gain is insufficient to offset the gains with the use of richer transformer embeddings, as shown in the next section.

\subsection{Modern NLP Methods}

All models described in this subsection have been fine-tuned using $10$-fold cross-validation, and the metric used for comparison is the average of weighted F1 scores across the folds. For the winning models, the fold model with the highest evaluation F1 score was further fine-tuned on the dataset for 2 additional epochs.

\subsubsection{Fine-tuning Language Models (LMs)}
% liu2019roberta,he2020deberta, beltagy2020longformer, tseng2023dynamicesg
We first fine-tune the XLM-RoBERTa (large) model \cite{liu2019roberta} using both the title and the main content of the news articles in each language. We also consider the Longformer (large-4096) \cite{beltagy2020longformer} model since some articles surpass the maximum token length of conventional BERT-style models\cite{kannan-seki-2023-textual}. Given the small size of the datasets per language, specifically Japanese, we also fine-tune multilingual models by combining the news articles in all four languages. Table \ref{tab:10foldcross_languagemodels} shows the results of the fine-tuning experiments. We did not fine-tune a monolingual model for Japanese due to the small training data size.

\begin{table}[!h]
    \centering
    \begin{tabular}[width = 0.5\textwidth]{p{4em}P{3em}cccc}
        \hline
        \textbf{Model-Lang} & \textbf{PSL} & \textbf{en} & \textbf{fr} & \textbf{kr} & \textbf{jp} \\ 
        \hline
        XLM-en  & \_ & 57.9 & \_ & \_ & \_ \\
        XLM-fr & \_ & \_ & 62.6 & \_ & \_ \\
        XLM-kr & \_ & \_ & \_ & \textbf{66.5} & \_ \\
        LF-en  & \_ & 57.7 & \_ & \_ & \_ \\
        LF-fr & \_ & \_ & \textbf{72.9} & \_ & \_ \\
        XLM-all & \_ & 57.4 & 71 & 66.1 & 56.7 \\
        LF-all & \_ & 50.2 & 57 & 39.8  & 39 \\ \hline 
        XLM-all  & direct & \textbf{60.2} & 71.3 & 63.4 & \textbf{59.9} \\
        LF-all & direct  & 46.6 & 55.2 & 44 & 44.1 \\ 
        LF-all & avg. conf. & 50.8 & 70.4 & 42.8 & 58.2 \\
        \hline
    \end{tabular}
    \caption{Weighted F1 score for impact duration classification averaged across 10-folds for a) fine-tuned LMs (rows 1-7), and b) semi-supervised learning (rows 8-10). Note: \textbf{XLM}: XLM-RoBERTa, \textbf{LF}: Longformer, \textbf{PSL}: Pseudo-label generation methods, \textbf{XLM-kr}: Korean\_Jetsons\_1 submission}
    \label{tab:10foldcross_languagemodels}
\end{table}

\subsubsection{Semi-supervised Learning}
%self-training
The training dataset is small and skewed across the impact duration labels. For English and French, 45-48\% of the articles belong to the `more than 5 years' class. For Korean, 55\% of the data belongs to the `less than 2 years' class. To overcome this class imbalance, we use a subset of the news articles released as part of an ESG issues classification task \cite{chen2023ESG}. 
% We use the finetuned multilingual Roberta and Longfromer models as teacher models and make predictions on English and French news articles in the Finnlp-2023 dataset to create pesudo-labels. 
We use the XLM and LF models in Table \ref{tab:10foldcross_languagemodels} as teacher models and make predictions on the English and French news articles in the ESG issues classification dataset. We generate pseudo labels in two ways: a) \textbf{direct}: use the label predicted by the multilingual teacher model directly, b) \textbf{avg. conf.}: for each article, take the average of the two confidence scores for each class predicted by the multi-lingual and mono-lingual teacher models and choose the label with the maximum average score.
We sample articles based on these pseudo labels and combine them with the original training data to reduce class imbalance. This augmented data is used to fine-tune XLM and LF models. The weighted F1 scores for these models are reported in Table \ref{tab:10foldcross_languagemodels}. The F1 scores of these models on the final test are reported in section \ref{sec:appendix-moden-nlp}.

\subsubsection{English Translation}

We also consider converting the problem from multi-lingual to mono-lingual by adding translation as a prerequisite for training and testing. We use the Google Translate API\footnote{https://translate.google.com/} to translate all non-English samples to English. Post translation, we fine-tune a DeBERTa-v3-xsmall model \cite{he2023debertav} on the class labels using both the article text and title (if available). The model experiment reports a 10-fold average weighted F1 score of 62.37 and a maximum weighted F1 of 66.82 on fold 8. The fold 8 model (DBERT-en) was used in the second submission (<lang>\_Jetsons\_2) for all languages.

\subsubsection{Ensemble}

The final model for the impact duration classification task is an ensemble model. We consider an ensemble of the three models - XLM, LF, and the DeBERTa-v3. The class label with the highest total model label score sum is used as the final class label. The submitted ensemble models were - \textbf{English\_Jetsons\_3} (XLM-all-direct, LF-all-avg. conf, DeBERTa-v3), \textbf{French\_Jetsons\_3} (XLM-all, LF-fr, DeBERTa-v3), \textbf{Korean\_Jetsons\_3} (XLM-kr, XLM-all, DeBERTa-v3), and \textbf{Japanese\_Jetsons\_3} (XLM-all-direct, DeBERTa-v3).

\section{Impact Level}
% We consider the impact level task for the English and French languages. 
We conduct experiments with the same two multilingual language models - XLM-RoBERTa-large and Longformer-large-4096 for the impact level task in French and English. We fine-tune the multilingual models in two ways: a) Using both languages, hoping that the data in one language can bolster the performance in the other, and b) separately in each language. First we compare models using only fold 0 data. Table \ref{tab:impact_level_results} shows that the weighted F1-score for the model trained in combined languages is lower than single language. So we use data in single language to further fine-tune the two models using all 10 folds of data and calculate the average results for each language. It shows that the XLM model has better performance in both languages: XLM-en (\textbf{65.02}) vs. LF-en (59.27), and XLM-fr (\textbf{65.29}) vs. LF-fr (63.84). We pick the XLM models with the best performance among 10 folds for each language as our first submission. As our second submission, we randomly chose a fold and used the best model fine-tuned on that fold.

\begin{table}[!htbp]
    \centering
    \begin{tabular}{lcccccc}
        \hline
        \textbf{La-} & \multicolumn{3}{c}{XLM} & \multicolumn{3}{c}{LF}\\
        \cline{2-7}
        \textbf{ng} & \textbf{en} & \textbf{fr} & \textbf{all} & \textbf{en} & \textbf{fr} & \textbf{all}\\ 
        \hline
        en & 57.3 & \_ & 51.7 & 56.1 & \_ & 47.6\\
        fr & \_ & 71.5 & 69 & \_ & 72.9 & 59.4\\
        \hline
    \end{tabular}
    % \caption{Weighted F1 score for impact level classification over fold 0 and all 10-folds respectively}
    \caption{Weighted F1 score for impact level classification over the data in the $0^{th}$ fold}
    \label{tab:impact_level_results}
\end{table}

\begin{table}[!htbp]
    \centering
    \begin{tabular}{lcc}
        \hline
        \textbf{Submission Model} & \textbf{Micro F1} & \textbf{Macro F1} \\ 
        \hline
        English\_Jetsons\_1 & 64.71 & \textbf{52.47} \\
        Korean\_Jetsons\_1 & \textbf{70} & \textbf{66.24} \\ 
        Japanese\_Jetsons\_2 & 36.5 & \textbf{25.6} \\
        French\_Jetsons\_1 & 47.95 & 37.06 \\
        \hline
        English\_Jetsons\_1 (IL) & \textbf{65.44} & \textbf{60.90} \\
        \hline
    \end{tabular}
    \caption{F1 scores on the test set, \textbf{Bold faced} ones are top on the leaderboard. IL indicates impact level}
    \label{tab:test_set_metrics}
\end{table}

\section{Analysis}

\begin{figure}[thb!]
    \centering
    \begin{subfigure}{0.49\linewidth}
        \centering
        \includegraphics[width=\textwidth]{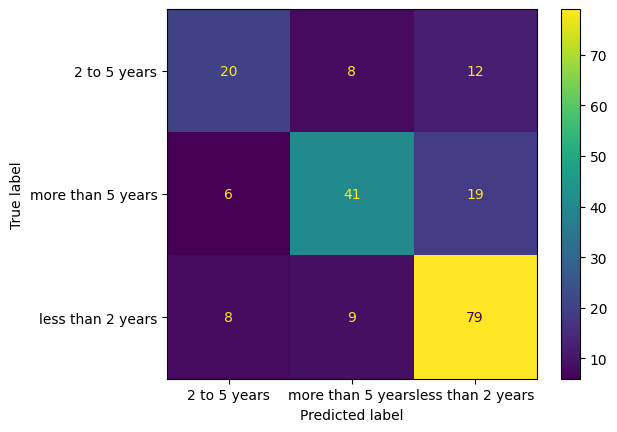}
        \caption{Korean\_Jetsons\_1}
        \label{fig:jetsons_korean_2}
    \end{subfigure}
    \hfill
    \begin{subfigure}{0.49\linewidth}
        \centering
        \includegraphics[width=\textwidth]{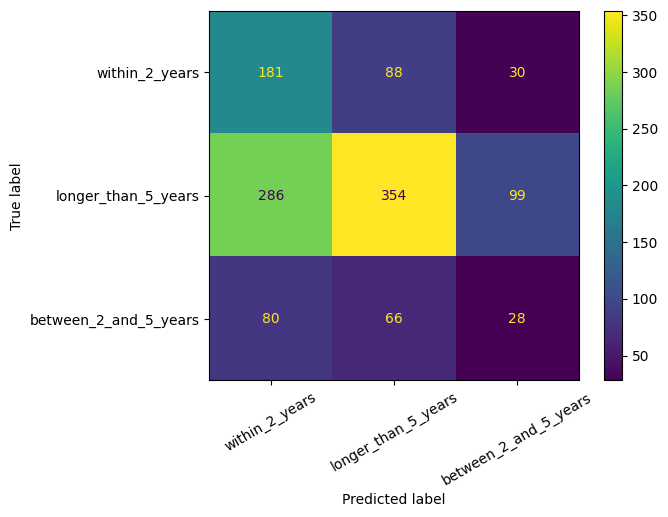}        
        \caption{Japanese\_Jetsons\_2}
        \label{fig:jetsons_japanese_2}
    \end{subfigure}
    \caption{ID confusion matrices}
    \label{fig:confusion_matrices_1}
\end{figure}

\begin{figure}[thb!]
    \centering
    \begin{subfigure}{0.49\linewidth}
        \centering
        \includegraphics[width=\textwidth]{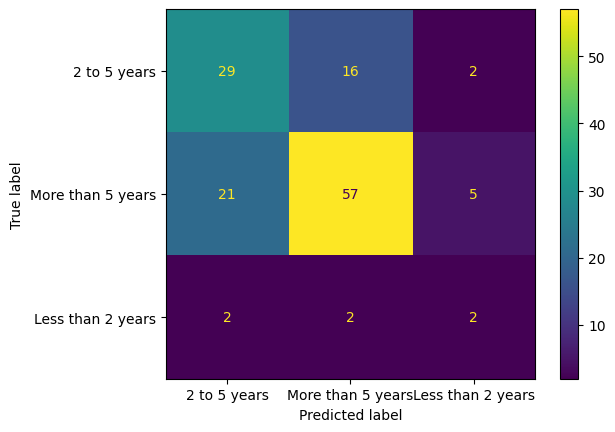}        
        \caption{English\_Jetsons\_3}
        \label{fig:jetsons_english_3}
    \end{subfigure}
    \hfill
    \begin{subfigure}{0.49\linewidth}
        \centering
        \includegraphics[width=\textwidth]{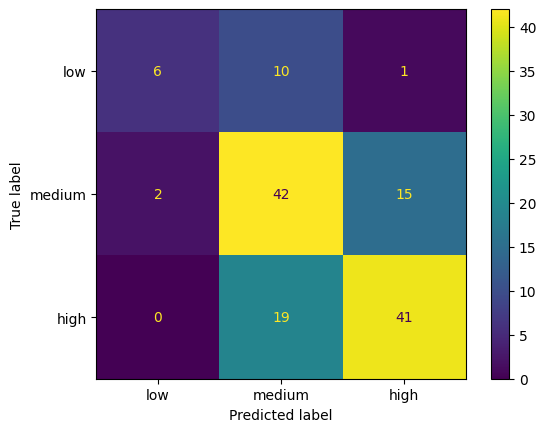}        
        \caption{English\_Jetsons\_1}
        \label{fig:jetsons_english_1_il}
    \end{subfigure}    
    \caption{Confusion matrices for: ID (a) and IL (b).}
    \label{fig:confusion_matrices_2}
\end{figure}

Table \ref{tab:test_set_metrics} shows the best micro and macro F1 scores on the test set for the submitted models. These models ranked best on 4 out of 7 tasks. Figures \ref{fig:confusion_matrices_1} and \ref{fig:confusion_matrices_2} show the confusion matrices for predictions generated using these four predictions. For impact duration, the models get most confused between `less than 2 year' and 'More than 5 years' classes. For impact level, `medium' and `high' are the most confusing classes.

\section{Conclusion}

ESG is increasingly important for stakeholders who want to align their values with their investments, reduce risks, and enhance long-term returns. For the FinNLP shared task of impact duration and level classification, we find that finetuning BERT-style models, along with data augmentation techniques like translation and self-training, perform the best. For impact duration in Korean and impact level in English, we find that fine-tuning a BERT-based classifier with a custom strategy performs the best. An ensemble with BERT-style models fine-tuned for impact duration in English using self-training and on just English translations performs best. The DeBERTa-v3 model fine-tuned on only English translations performs best on the Japanese dataset for the impact duration task.

\section{Bibliographical References}\label{sec:reference}

\bibliographystyle{lrec-coling2024-natbib}
\bibliography{lrec-coling2024-example}

\appendix

\section{Additional experiment details}

\subsection{Traditional NLP methods}
\begin{figure*}[t]
    \centering
    \includegraphics[width=\textwidth]{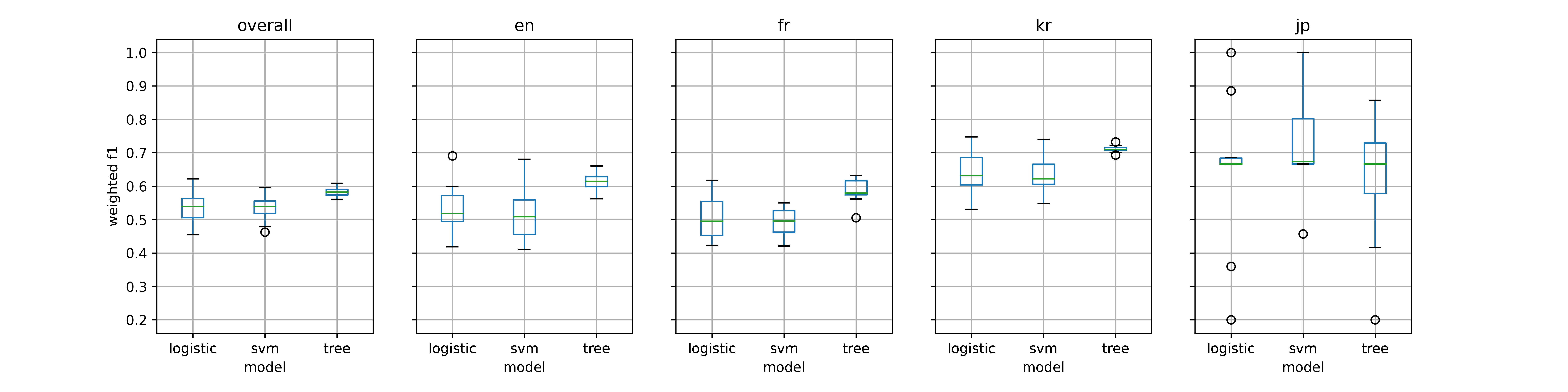}
    \caption{Weighted F1 scores across 10-folds for traditional NLP models for predicting impact length}
    \label{fig:baselinef1}
\end{figure*}

The weighted F1 score across different models and different languages is summarized in Figure \ref{fig:baselinef1}.

\subsection{Modern NLP Methods}\label{sec:appendix-moden-nlp}
\begin{figure}[t]
\centering
    \includegraphics[width=\linewidth]{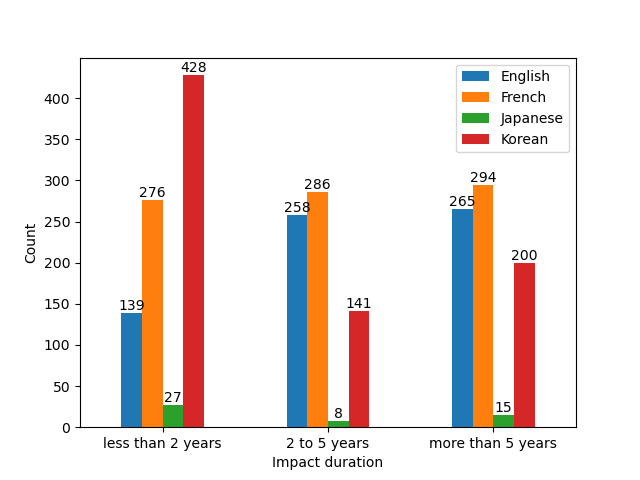}
        \caption{Data distribution across classes in the training set used in semi-supervised learning} 
        \label{fig:semi_supervised_training}
\end{figure}

% TODO for camera ready: add label distribution for self-training model, bold the best models on test set in table
The distribution of data across class labels in the dataset used to train the student models using semi-supervised learning is shown in figure \ref{fig:semi_supervised_training}. The micro and macro F1 scores achieved by the different fine-tuned language models on the test set are reported in table \ref{tab:self-training-test-set}. In the case of the English, Korean, and Japanese data, we see that the models with the best 10-fold cross-validation scores also perform similarly on the test set. However, for the French news articles, while fine-tuning the Longformer model using only French data (LF-fr) gives maximum average weighted F1 during cross-validation, the same isn't reflected on the test set. XLM-Roberta fine-tuned on articles in all languages along with self-training (row 8 in table \ref{tab:self-training-test-set}) gives the best macro F1 of \textbf{50.54} and micro F1 of \textbf{53.42}. The scores on the Japanese test data have been calculated after removing the 327 unlabelled news articles.

\begin{table*}[t]
\centering
% \resizebox{\columnwidth}{!}{%
\begin{tabular}{p{4em}P{4em}cccccccc}
\hline
\textbf{Model-lang} & \textbf{PSL}                                          & \multicolumn{2}{c}{\textbf{en}} & \multicolumn{2}{c}{\textbf{fr}} & \multicolumn{2}{c}{\textbf{kr}} & \multicolumn{2}{c}{\textbf{jp}} \\ \hline
                                                               &                                                       & Mi. F1       & Ma. F1       & Mi. F1       & Ma. F1       & Mi. F1       & Ma. F1       & Mi. F1       & Ma. F1       \\ \hline
XLM-en                                                         &                                                       & \textbf{61.03}          & \textbf{48.79}          &                &                &                &                &                &                \\
XLM-fr                                                         &                                                       &                &                & 54.79          & 43.01          &                &                &                &                \\
XLM-kr                                                         &                                                       &                &                &                &                & \textbf{70.0}           & \textbf{66.24}          &                &                \\
LF-en                                                          &                                                       & 55.88          & 44.9           &                &                &                &                &                &                \\
LF-fr                                                          &                                                       &                &                & 47.95          & 37.06          &                &                &                &                \\
XLM-all                                                        &                                                       & 58.82          & 43.66          & 47.95          & 43.28          & 64.0           & 57.09          & \textbf{38.19}          & \textbf{32.99}          \\
LF-all                                                         &                                                       & 59.56          & 38.3           & 43.84          & 36.86          & 48             & 21.62          & 24.81           & 13.25           \\ \hline
XLM-all                                                        & direct                                                & 61.03          & 46.7           & \textbf{53.42}          & \textbf{50.54}          & 67             & 62.64          & 36.23         & 31.97          \\
LF-all                                                         & direct                                                & 56.62          & 47.11          & 42.47          & 35.87          & 50.5           & 43.62          & 24.81           & 13.25           \\
LF-all                                                         & avg. conf. & 58.82          & 45.16          & 38.36          & 34.34          & 47.5           & 27.36          & 28.3           & 18.26  \\ \hline         
\end{tabular}%
% }
\caption{Micro and Macro F1 for impact length classification task in the final test set.}
\label{tab:self-training-test-set}
\end{table*}

\section{Hyperparameters}

For the XLM-RoBERTa and Longformer fine-tuning experiments, the learning rates for the mono-lingual and multi-lingual models were $2e-5$ and $8e-6$, respectively, along with batch size -8 and epochs - 10. The Longformer-large models were fine-tuned with gradient accumulation of 2 steps. For the DeBERTa-v3-xsmall model, the following hyperparameters were: learning rate - $2e-05$, epochs - 10, weight decay $0.01$, and batch size - 2. The fine-tuning process was carried out on a GPU with 32 GB memory.

\end{document}